\documentclass[9pt,twocolumn,twoside]{pnas-new}

\usepackage{enumitem} 
\usepackage{hyperref}
\usepackage{url}
\usepackage{algorithm}
\usepackage{subfiles}
\usepackage{amsmath}
\usepackage{amssymb}
\usepackage{mathtools}
\usepackage{amsthm}
\usepackage{microtype}
\usepackage{graphicx}
\usepackage{booktabs}
\usepackage{hyperref}
\usepackage{natbib}
\usepackage{xcolor}
\usepackage{amsfonts}

\newtheorem{theorem}{Theorem}

\theoremstyle{definition}

\theoremstyle{remark}

\usepackage{times}
\usepackage{latexsym}
\usepackage[T1]{fontenc}
\usepackage[utf8]{inputenc}
\usepackage{microtype}
\usepackage{inconsolata}
\usepackage{graphicx}
\usepackage{amsmath}
\usepackage{enumitem}
\usepackage{graphicx}    
\usepackage{subcaption}  
\usepackage{caption}
\usepackage[utf8]{inputenc}
\usepackage{CJKutf8}
\usepackage{wrapfig} 
\usepackage{tikz}
\usetikzlibrary{shapes.geometric, arrows}
\usepackage[utf8]{inputenc} 
\usepackage[T1]{fontenc}    
\usepackage{hyperref}       
\usepackage{url}            
\usepackage{booktabs}       
\usepackage{amsfonts}       
\usepackage{nicefrac}       
\usepackage{microtype}      
\usepackage{xcolor}         
\templatetype{pnasresearcharticle} 

\DeclareMathOperator{\Mis}{Mis}

\begin{document}

\title{Hallucination, Monofacts, and Miscalibration: An Empirical Investigation}

\author[a, 1]{Miranda Muqing Miao}
\author[a, 1]{Michael Kearns}

\affil[a]{Department of Computer and Information Science, University of Pennsylvania, 3333 Chestnut Street, Amy Gutmann Hall, Philadelphia, PA 19104, USA}

\leadauthor{Miao}

\significancestatement{
We show that hallucination in large language models can be controlled through deliberate manipulation of training data frequency distributions. By sampling training data from heavy-tailed distributions that naturally reduce rare facts, and by strategically repeating small subsets of training examples during fine-tuning, we reduce hallucination rates by up to 40\% without sacrificing accuracy. Our findings challenge the widespread practice of deduplicating training data and reveal that the distribution of fact frequencies could fundamentally affect model reliability. This work establishes training data composition as a primary lever for hallucination control, offering practitioners a simple, interpretable alternative to complex post-hoc intervention methods that operate on model internals.
}

\authorcontributions{Author contributions: M.M.M. and M.K. designed research; performed research; analyzed data; and wrote the paper.
\\Reviewers: A.K., OpenAl; and S.S.V., Georgia Institute of Technology.}

\authordeclaration{The authors declare no competing interests.}

\correspondingauthor{\textsuperscript{1}To whom correspondence may be addressed. Email: miaom@seas.upenn.edu, mkearns@cis.upenn.edu
\\ Published February 19, 2026.}


\begin{abstract}
Hallucinated facts in large language models (LLMs) have recently been shown to obey a statistical lower bound determined by the \emph{monofact rate} (related to the classical Good-Turing missing mass estimator) minus model miscalibration [A. T. Kalai,
S. S. Vempala, "Calibrated language models must hallucinate" in Proceedings of the 56th Annual ACM Symposium on Theory of Computing (STOC) (New York, NY, USA, 2024), pp. 160–171]. We present the first empirical investigation of this three-way relationship in classical $n$-gram models and fine-tuned transformer models. By generating training data from Pareto distributions with varying shape parameters, we systematically control the monofact rate and establish its positive relationship with hallucination. To bridge theory and practice, we derive an empirical analog of the hallucination bound by replacing the population miscalibration term (Section 1.1) with an \emph{empirical} bin-wise KL divergence and confirm its practical viability. We then introduce selective upweighting --- a simple yet effective technique that strategically repeats as little as 5\% of training examples --- to deliberately \emph{inject miscalibration} into the model. This intervention reduces hallucination by up to 40\%, challenging universal deduplication policies. Our experiments reveal a critical trade-off: selective upweighting maintains pre-injection levels of accuracy while substantially reducing hallucination, whereas standard training gradually improves accuracy but fails to address persistently high hallucination, indicating an inherent tension in optimization objectives. 
\end{abstract}

\doi{\url{https://doi.org/10.1073/pnas.2533582123}}

\maketitle
\thispagestyle{firststyle}
\ifthenelse{\boolean{shortarticle}}{\ifthenelse{\boolean{singlecolumn}}{\abscontentformatted}
{\abscontent}}{}

\firstpage{5}

\label{sec:introduction}
    \dropcap{C}onsider the case of writing a biography of a living person with a large language model (LLM). The model may confidently state that ``John Smith was born in Seattle in 1982, earned his PhD from Stanford in 2008, and now leads AI research at Tech Corp,'' but these ``facts'' could be fabricated. Such hallucinations (plausible but verifiably false statements) are a critical issue for language models, especially in high stakes scenarios: when lawyers submit hallucinated legal cases or doctors receive incorrect medical advice \cite{pham2024towards, shin2023humiliated}.

A growing body of work traces hallucination back to how models encode, retain, and forget factual information during pre-training and supervised fine‑tuning (SFT). During pre-training, synthetic recall tasks reveal learning plateaus and inaccuracy spikes when models learn new facts \cite{zucchet2025learn}. During SFT, exposure to unfamiliar data causes an increase in hallucination, underscoring how new knowledge acquisition can compromise generation fidelity~\cite{kang2024unfamiliar}. Other studies introduce methods such as embedding non‑parametric memory into the learning objective and injecting event memories directly into the weights of LLMs \cite{zhong2022memory, pan2025memorization}. 

A second research direction develops post-hoc hallucination reduction techniques that are applied after pre-training and SFT. Some find that latent‑space steering via a truthfulness separator vector can create separable clusters of faithful and fabricated outputs \cite{park2025steer}. Others realize that follow‑up questioning can expose inconsistencies in fabricated citations, allowing models to self‑diagnose hallucinations \cite{agrawal2023hallucinating}. Other work finds ways to suppress attention layers that disproportionately drive fabrication \cite{li2024look}. While effective, these intervention methods address symptoms rather than the fundamental statistical mechanisms that cause hallucinations.

Recent theory by Kalai and Vempala  provides a formal and theoretical treatment of hallucination~\cite{kalai2024calibrated}. Their work demonstrates a more fundamental cause: calibrated language models must hallucinate at a rate associated with the prevalence of rare facts (related to the classical Good-Turing missing mass estimator) in their training data. Specifically, they prove that for arbitrary facts whose veracity cannot be systematically determined from training data, their hallucination rate has a statistical lower bound tied to the fraction of facts that appear exactly once in training (the \emph{monofact} rate) minus model miscalibration. Model miscalibration is the sum of bin-wise absolute difference between the model’s predicted confidence scores and the corresponding probability mass under the true data distribution (see full definition in Section 1.1). 

\newpage
In our work, we restrict attention to \emph{factual hallucinations} to align with the Kalai-Vempala framework by focusing on attribute-level correctness in generated statements (e.g., a movie's actor or title; a person’s birth‑year or job title). 

We demonstrate that the statistical mechanisms governing hallucination in language models can be understood and manipulated through deliberate control of training data properties. We test whether the theoretical relationship between monofact prevalence, model miscalibration, and hallucination identified by Kalai and Vempala~\cite{kalai2024calibrated} can guide practical mitigation strategies across model architectures and scales.

We validate this framework empirically by confirming that monofact rate positively correlates with hallucination rates in both $n$-gram models and fine-tuned large language models. Crucially, we show that intentional miscalibration could reduce hallucination by up to 40\% while holding monofact rates constant. We test these relationships using structured movie facts (comma-separated tuples) for $n$-gram models and naturalistic biographical text for supervised fine-tuning.

To enable practical application, we provide an empirical analog of the population-level miscalibration term in the Kalai-Vempala theorem with empirical KL-divergence, which requires no knowledge of the true data distribution. We generate training data using heavy-tailed Pareto distributions to satisfy the i.i.d.\ sampling requirements of the theoretical framework while reflecting established power-law patterns in natural language~\cite{piantadosi2014zipf,ferrer2001two,baayen2001word}. Among several distribution families tested, Pareto distributions prove most effective for controlling monofact rates.

Extending recent work showing that power-law-distributed pretraining data accelerates factual recall~\cite{zucchet2025learn}, we systematically map Pareto shape parameters to monofact rates during supervised fine-tuning. We find that lower monofact rates consistently reduce hallucination and shorten learning plateaus. We introduce a practical intervention: sample upweighting during training as a way to inject higher miscalibration into a trained model. Applied to a small sample of the training examples, this approach reduces hallucination significantly in 6 of 8 configurations tested ($p < 0.01$), with effectiveness varying systematically by monofact rate.

\newcommand{\Yset}{\mathcal{Y}}        
\newcommand{\Fset}{\mathcal{F}}        
\newcommand{\Hset}{\mathcal{H}}        
\newcommand{\Oset}{\mathcal{O}}        
\newcommand{\Uset}{\mathcal{U}}        
\newcommand{\Xset}{\mathcal{X}}        
\newcommand{\MF}{\text{MF}}                  

\section{Preliminaries}
\phantomsection
\label{sec:theory_prelim}
To investigate hallucination in a controlled setting, we need to carefully define what constitutes a fact versus a hallucination. Consider the statement ``Timothée Chalamet starred in Dune: Part Two directed by Denis Villeneuve.'' This is a true movie fact. However, if our model generates ``Timothée Chalamet starred in Dune: Part Two directed by Christopher Nolan,'' this would be a hallucination --- a plausible but false statement.

Let \(U\) be the universe of all plausible statements (or ``factoids'') in a given domain. We partition $U$ into $T$, the subset of true statements, and $F = U \setminus T$, 
the complementary set of falsehoods. The labels $T$ and $F$ represent ground-truth annotations rather than model predictions. During training, the model sees only training data \(S\subset T\); the unobserved remainder is \(H=U\setminus S\).\footnote{Notation differs from \citealp{kalai2024calibrated}: their \(Y,F,H,O,U\) correspond to our \(U,T,F,S,H\).} After training, the language model assigns a distribution \(g\) over $T$ and generates a sample set \(G\). The truth set \(T\) is governed by the true distribution \(p\), which we control experimentally. 

\begin{theorem}[Kalai and Vempala's Hallucination Lower Bound]
\begin{equation}
\label{eq:kalai_theorem}
  f_{gen}\;\ge\;
  \hat{\text{MF}}
  -\Mis(g,p)
  -\frac{3e^{-m}}{\delta}
  -\sqrt{\tfrac{6\ln(6/\delta)}{n}}
\end{equation}
\end{theorem}
\noindent
where \(\hat{\text{MF}}\) is the empirical \emph{monofact} rate (fraction of facts seen exactly once in \(S\)), \(\Mis(g,p)\) is the miscalibration between \(g\) and true \(p\) (see definition below), \(m\) is a sparsity parameter, \(n=|S|\), and \(\delta\) is the confidence level. For realistic data sizes the final two terms are negligible, leaving hallucination essentially governed by monofact rate and miscalibration. See Kalai-Vempala for details.

\smallskip
While treating the Kalai-Vempala bound as an equality to derive precise coefficients for monofact rates and miscalibration seems appealing, our explorations showed these coefficients fluctuate significantly across model class, model capacities, and data characteristics. We therefore focus on the qualitative three-way relationship, yielding qualitative insights for practical language model training.

Rather than strictly checking ~\eqref{eq:kalai_theorem}, we treat it as a guide and vary \(\hat{\text{MF}}\) and \(\Mis(g,p)\) directly to study their practical impact on hallucination. Key assumptions for the framework are (i) i.i.d.\ sampling from \(p\), (ii) a sparse truth set relative to \(U\), (iii) \(S\) only samples true facts from \(T\). Additional assumptions can be found in SI Appendix.

We follow Kalai and Vempala's definition of hallucination \cite{kalai2024calibrated}. A generation is a hallucination if it lies outside the truth set \(T\), i.e.\ \(F_{\text{gen}} \coloneqq G \cap F\); the hallucination rate is then \(f_{\text{gen}} = \lvert F_{\text{gen}}\rvert / \lvert G\rvert\).

Kalai et al. (2025) generalized the lower bound in ~\eqref{eq:kalai_theorem} using a reduction from Is-It-Valid (IIV) classification, recovering it as a special result for arbitrary-fact hallucinations in their Section~3.3.1 ~\cite{kalai2025}. This 2025 formulation shows that hallucination is lower-bounded by twice the IIV misclassification rate, minus calibration terms. Our experimental design aligns with Section~3.3.1 of Kalai et al. ~\cite{kalai2025}. However, We focus on the monofact-calibration framework because these quantities can be directly observed and manipulated through training data structure, whereas IIV classification rates are less amenable to data-centric intervention.

\subsection{Calibration}
\label{sec:miscalibration}

Calibration asks whether probabilities assigned by a model match empirical frequencies.  
Analogous to a weather forecaster whose ``30\% rain’’ on x days prediction should coincide with rain on roughly 30\% of those days, a language model is calibrated if, among statements it rates near \(q\), the proportion that are factual is also \(q\).  


\paragraph{Logarithmic binning.}
Following Kalai and Vempala~\cite{kalai2024calibrated}, we partition the unit interval into logarithmic bins and set $\epsilon = 0.1$ unless otherwise indicated. 
\[
B_i=[(1-\epsilon)^{\,i+1},\,(1-\epsilon)^{\,i}],\quad i\ge 0
\]
and place each statement \(x\) in the unique bin whose range contains \(g(x)\). We have also tested adaptive partitions, yielding indistinguishable results.\footnote{Let \(b\in\mathbb{N}\) and define thresholds \(a_i\) by
\(\sum_{x:\,g(x)\le a_i} g(x)=i/b\); the partition \(\{(0,a_1],\,(a_1,a_2],\ldots\}\) balances mass across \(b\) bins.}

\paragraph{Miscalibration metric.}
For any partition \(\mathcal{B}(g)\) induced by \(g\), form the coarsened distribution
\(p^{\mathcal{B}(g)}(x)=p(B)/|B|\) for \(x\in B\). Miscalibration is the total‑variation distance between \(g\) and this coarsening.
Then, miscalibration is defined as:

\begin{align} 
\text{Mis}(g, p) &\coloneqq \|p^{\mathcal{B}(g)} - g\|_{\text{TV}} \label{eq:mis_gen} \\
&= \frac{1}{2} \sum_{B \in \mathcal{B}(g)} \sum_{x \in B} \left| \frac{p(B)}{|B|} - g(x) \right| \label{eq:mis}
\end{align}

Because \(p(B)=\sum_{x\in B}p(x)\) and \(g(B)=\sum_{x\in B}g(x)\), ~\eqref{eq:mis_gen} is equivalently the sum of within‑bin total‑variation distances,
\[
\textstyle\Mis(g,p)=\sum_{B\in\mathcal{B}(g)}\mathrm{TV}_{B}(p,g),
\]
with \emph{larger values} indicating worse calibration.Throughout this paper miscalibration is measured exclusively on the pair \((g,p)\) under the partition defined by \(g\).

While minimal miscalibration or so-called perfect calibration is typically desirable in machine learning, the Kalai-Vempala framework reveals a tension: perfect calibration on rare facts could force models to assign probability to unseen (potentially false) completions. Our intervention deliberately introduces miscalibration to concentrate probability mass on well-learned facts, reducing hallucination. In this context, strategically directed miscalibration is beneficial.

\subsection{An Empirical Analog}
\label{sec:new_kl_theorem}

Kalai–Vempala’s bound includes miscalibration, a term that
depends on the unknown truth distribution \(p\), which is often unobserved in real-world applications.
We replace this unobservable term with the \emph{empirical} bin-wise
Kullback–Leibler divergence
\[
  D_{\mathrm{KL}}\!\bigl(\hat p_{\mathcal{B}(g)}\Vert g_{\mathcal{B}(g)}\bigr)
  \;=\;
  \sum_{B\in\mathcal{B}(g)}\hat p_B\,\log\!\frac{\hat p_B}{g_B}
\]
where \(\mathcal{B}(g)\) is the probability‑bin partition defined in
Section 1.1,
\(\hat p_B\) is the empirical frequency of true statements in bin \(B\)
measured on the training sample \(S\),
and \(g_B\) is the model’s total probability mass on that bin.

We adopt KL divergence to address the empirical-model gap because KL divergence is similar to cross-entropy loss, the standard training objective for modern language models. This choice provides a natural and interpretable metric that directly measures how the model's learned  distribution deviates from empirical frequencies in the same units used during optimization. An analogous bound using total variation or empirical Expected Calibration Error (ECE) follows from similar mathematical arguments; However, we present the KL formulation for its closer connection to training dynamics.

Throughout this paper, all reported miscalibration values are computed using the exact definition of~\eqref{eq:mis_gen}, following Kalai and Vempala (2024)~\cite{kalai2024calibrated}.

The empirical KL-divergence $D_{\mathrm{KL}}(\hat{p}_{B(g)} \| g_{B(g)})$ serves as a \emph{practical proxy} when the true distribution $p$ is unavailable; in our controlled experiments where $p$ is known, we report both metrics. 

\smallskip
\begin{theorem}[Empirical KL‑Divergence Hallucination Bound]
Fix $b = \lvert\mathcal{B}(g)\rvert>1$, sparsity parameter $m$ as in ~\eqref{eq:kalai_theorem}, and confidence $\delta\in(0,1)$.
With probability at least \(1-\delta\) over the draw of \(S\),
\begin{multline}
\label{eq:kl_theorem}
  f_{gen}
  \;\ge\;
  \hat{\mathrm{MF}}
  \;-\;\sqrt{\tfrac12\,D_{\mathrm{KL}}\!\bigl(\hat p_{\mathcal{B}(g)}\Vert g_{\mathcal{B}(g)}\bigr)} \\
  -\;\frac{3e^{-m}}{\delta}
  \;-\;\sqrt{\frac{6\ln(12/\delta)}{n}}
  \;-\;\sqrt{\frac{b\ln 2 + \ln(2/\delta)}{2n}}.
\end{multline}
\end{theorem}

\paragraph{Proof sketch.}
A triangle inequality splits \(\Mis(g,p)\) into (i) the \emph{sampling} error \(\|p_{\mathcal{B}(g)}-\hat p_{\mathcal{B}(g)}\|_{\mathrm{TV}}\) and (ii) the \emph{empirical–model} gap \(\|\hat p_{\mathcal{B}(g)}-g_{\mathcal{B}(g)}\|_{\mathrm{TV}}\). The Bretagnolle--Huber--Carol inequality controls (i) by \(O\!\bigl(\sqrt{b/n}\bigr)\), while Pinsker's inequality turns (ii) into \(\sqrt{\tfrac12 D_{\mathrm{KL}}(\hat p_{\mathcal{B}(g)}\Vert g_{\mathcal{B}(g)})}\). A union bound on these two events delivers inequality~\eqref{eq:kl_theorem}; see SI Appendix for details.

\paragraph{Intuition.} While miscalibration is measured exclusively on \((g,p)\), \(D_{\mathrm{KL}}(\hat p_{\mathcal{B}(g)}\Vert g_{\mathcal{B}(g)})\) in inequality~\eqref{eq:kl_theorem} is entirely data‑driven and measured on \((g,\hat p)\). With sufficiently large \(n\), as \(D_{\mathrm{KL}}(\hat p_{\mathcal{B}(g)}\Vert g_{\mathcal{B}(g)})\!\to\!0\), the lower bound approaches the empirical monofact rate \(\hat{\text{MF}}\). The additional sampling penalty
\(\sqrt{(b\ln 2 + \ln(2/\delta))/(2n)}\) requires that the number of bins \(b\) be small relative to \(n\):
because the term scales as \(O(\sqrt{b/n})\), one needs \(n \gg b\) for the penalty to vanish\footnote{For instance, choosing a bin width \(\varepsilon = 0.1\) yields \(b \approx 200\) bins. Choosing \(n = 10^{5}\) with $\delta$ = 0.01, the term is approximately 0.027.}.    
\subsection{Constructing Data from Pareto Distributions}\label{sec:dataset}

\begin{figure*}[!t]  
    \centering
    \includegraphics[width=0.8\textwidth]{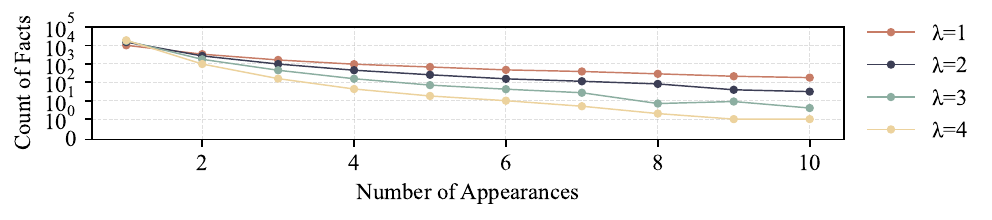}  
    \captionsetup{skip=4pt}
    \caption{Illustration of fact repetition frequencies across varying parameters ($\gamma$). Plot shows how often a fact appears
(x-axis) versus how many facts have that appearance count (y-axis, logarithmic scale).}
    \label{fig:sample_distr}  
\end{figure*}

To construct $p$ and vary the monofact rate we replicate each fact in \(T\) a random number of times, drawing that count from several candidate distributions. Gaussian and Poisson choices compress mass toward the mean and fail to generate a diverse range of monofact rates (see Fig.~S4 in SI Appendix). A heavy‑tailed Pareto distribution
\[
  f(x;\gamma,x_m)=\frac{\gamma x_m^{\gamma}}{x^{\gamma+1}},\qquad x\ge x_m,
\]
with shape \(\gamma\) and scale \(x_m\), proves ideal. Sweeping \(\gamma\) with a fixed \(x_m\) = 1 yields a smooth continuum from dense duplication to monofact‑dominated samples (Fig.~\ref{fig:sample_distr}). The resulting multiset constitutes our population distribution \(p\). We sample with replacement from the expanded multiset to form \(S\), a subset of \(T\). Adjusting \(\gamma\) thus directly tunes the empirical monofact rate \(\hat{\text{MF}}\) while preserving i.i.d.\ sampling.

\section{$n$‑Gram Methodology}
\label{sec:ngram_method}

We start with classical $n$‑gram models because they serve as an ideal controlled environment where we have authority over experimental variables including data generation, model architecture, training process, and evaluation methods. This provides a laboratory setting to test the Kalai-Vempala theoretical framework before advancing to more complex LLM experiments. These $n$‑gram models offer an analytically tractable sandbox where we can produce thousands of trained models, unlike the compute-constrained LLM SFT experiments. 

To closely mirror the “Factoid Assumptions’’ of Kalai and Vempala~\cite[§4.1]{kalai2024calibrated}, our $n$-gram dataset consists of comma‑separated six‑tuples and each document only consists of one fact. Among several orders of $n$ we find \emph{bigrams} strike the best balance: they accommodate sufficient but not excessive local dependencies in our tuples to make hallucination observable and easily analyzed. Higher‑order models unsurprisingly memorize better and hallucinate less, but we observe that the quantitative relationship among monofact rate, miscalibration, and hallucination remains invariant across $n$.

For the $n$-gram model only, we frame ``facts’’ as structured, comma‑separated six‑tuples to focus a simplified language-setting without noise. We derive the true‑fact set \(T\) from the IMDb non‑commercial dataset \citep{imdb2024datasets}. Each entry is a six‑tuple \textit{Actor, Co‑star, Movie, Director, Genre, Year}. We sample \(10{,}000\) instances with replacement to train each $n$-gram model. 

We report results for bigram models; trigram experiments yielded qualitatively identical 
findings but risk being considered overpowered for our simple six-entity statements. Our design aligns with the prompt-free setting of Kalai and Vempala  (2024)~\cite{kalai2024calibrated}. For the SFT experiments, we additionally measure
prompt completion accuracy to capture performance on both prompt-free generation and cloze-like completion.

\subsection{Statement Generation Process}

Let \(g\) denote the distribution over six‑tuples
\(x=(t_1,\dots,t_6)\). For any pair of tokens \((t_1,t_2)\), the bigram model learns \(g(t_1,t_2)=g(t_1)\,g(t_2\mid t_1)\), where \(g(t_1)\) is the marginal over first tokens and \(g(t_2\mid t_1)\) the conditional over second tokens.  Generating a new six‑tuple proceeds sequentially, giving \(g(x)=g(t_1)\prod_{i=1}^{5} g(t_{i+1}\mid t_i)\).

To be more specific, we first draw \(t_1 \sim g(t_1)\) and then, for each position \(i = 1{:}5\), sample \(t_{i+1} \sim g\!\bigl(t_{i+1}\mid t_i\bigr)\). Note that while this approach is similar to standard bigram language modeling, it does not exploit the full structure of our movie data where each position has its own distinct token space. Instead, we treat all position pairs equivalently in building our transition probabilities. For evaluation, each model is trained on $|S| = 10,000$ statements and generates an equal number of new statements.

\subsection{Controlled Miscalibration Injection}
\label{sec:upweight}
To dissect the role of miscalibration while \emph{holding the monofact rate fixed}, we introduce a lightweight \textit{upweighting} routine that perturbs the learned transition counts of a trained bigram model to inject miscalibration into $g$.

Let the initial bigram counts be \(C_{0}(t_{i},t_{j})\).  
Choose a subset \(E_{k}\subset S\) of size \(k\).  
For every token pair \((t_{i},t_{j})\) occurring in \(E_{k}\), add one extra count:  
\(C_{1}(t_{i},t_{j}) \leftarrow C_{0}(t_{i},t_{j}) + \mathbf{1}_{(t_{i},t_{j})\in E_{k}}\).  
The same increment is applied to initial‑state counts.  
Fig.~S1 in the SI Appendix details this update with comprehensive steps and then calls the normalization subroutine to convert counts back into normalized probabilities.

Upweighting only a fraction of the training data induces selective over‑confidence, while keeping the empirical monofact rate constant. As \(k\) grows, miscalibration and hallucination first increase and decline to baseline once \(E_{k}=S\) (the global duplication cancels after normalization). This controlled knob lets us chart hallucination as a function of miscalibration.

\section{$n$‑Gram Model Results}
\label{sec:ngram_results}


\subsection{Monofact and Hallucination}
\label{sec:pareto_results}
\begin{figure*}[!tttt]  
    \centering
    \includegraphics[width=0.8\textwidth]{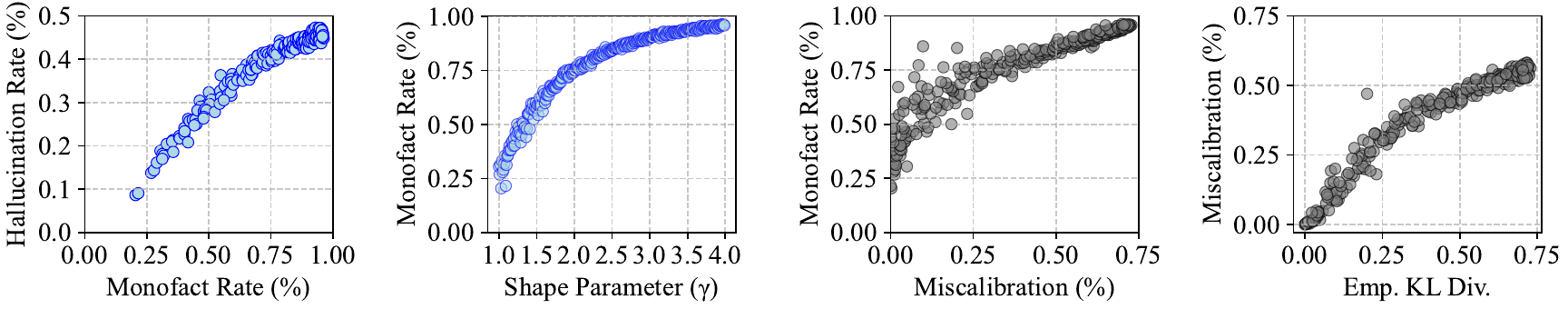}  
    \captionsetup{skip=4pt}
    \caption{Each dot represents a sample of statements. \textbf{Left:} Results show a positive relationship between monofact rate and hallucination. \textbf{Middle Left:} Heavy-tailed document distributions yield lower monofact rates. \textbf{Middle Right:} Miscalibration increases with monofact rate, suggesting better learning and calibration in low-monofact distributions. \textbf{Right:} Strong positive correlation between empirical KL divergence and miscalibration metrics.}
    \label{fig:pareto}  
\end{figure*}

Our controlled simulations validate the monofact-hallucination relationship in~\eqref{eq:kalai_theorem}. The left panel of Fig.~\ref{fig:pareto} shows an approximately linear rise in hallucination as the monofact rate grows: the hallucination rate climbs from \(\approx0\%\) to \(\approx50\%\) when the monofact share increases from \(0\%\) to \(100\%\). This trend roughly matches the factor‑\(0.5\) slope.

The middle right panel of Fig.~\ref{fig:pareto} confirms that monofact rate and miscalibration naturally move in tandem without intervention. The right panel further shows that empirical KL divergence successfully tracks miscalibration, supporting its use as a practical proxy for our defined setting.

\begin{figure*}[!t]  
    \centering
    \includegraphics[width=0.75\textwidth]{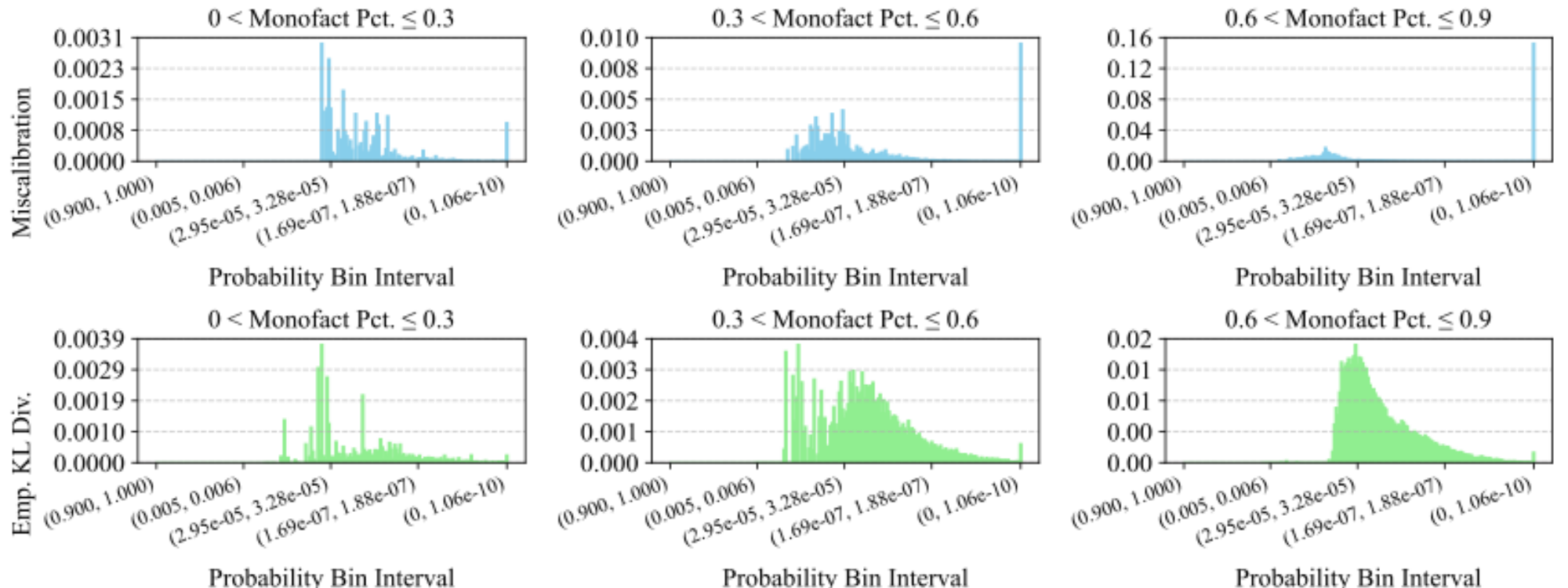} 
    \captionsetup{skip=4pt}
    \caption{\textbf{Top:} Average miscalibration per probability bin (binning created by a logarithmic binning strategy with $\epsilon$ = 0.1) for three monofact percent ranges. \textbf{Bottom:} Average empirical KL divergence per probability bin for three monofact percent ranges.
}
    \label{fig:ngram_upweighting_bin}  
\end{figure*}

Fig.~\ref{fig:ngram_upweighting_bin} reports mean bin‑wise miscalibration for three monofact ranges \emph{without any miscalibration injection}. As monofact rates and hallucination fall, miscalibration in the right‑most (lowest‑confidence) bin contracts towards zero and we observe a similar increase in \emph{polarity}. What dampens hallucination organically is not simply the magnitude of miscalibration, but relative polarity of bin-wise miscalibration gained from sampling training data using Pareto distributions. 

\subsection{Miscalibration Injection as a Hallucination‑Reduction Mechanism}
\label{sec:ngram_miscal_injection}
\begin{figure*}[h]  
    \centering
    \includegraphics[width=0.8\textwidth, trim = {4mm 2mm 4mm 2mm}, clip]{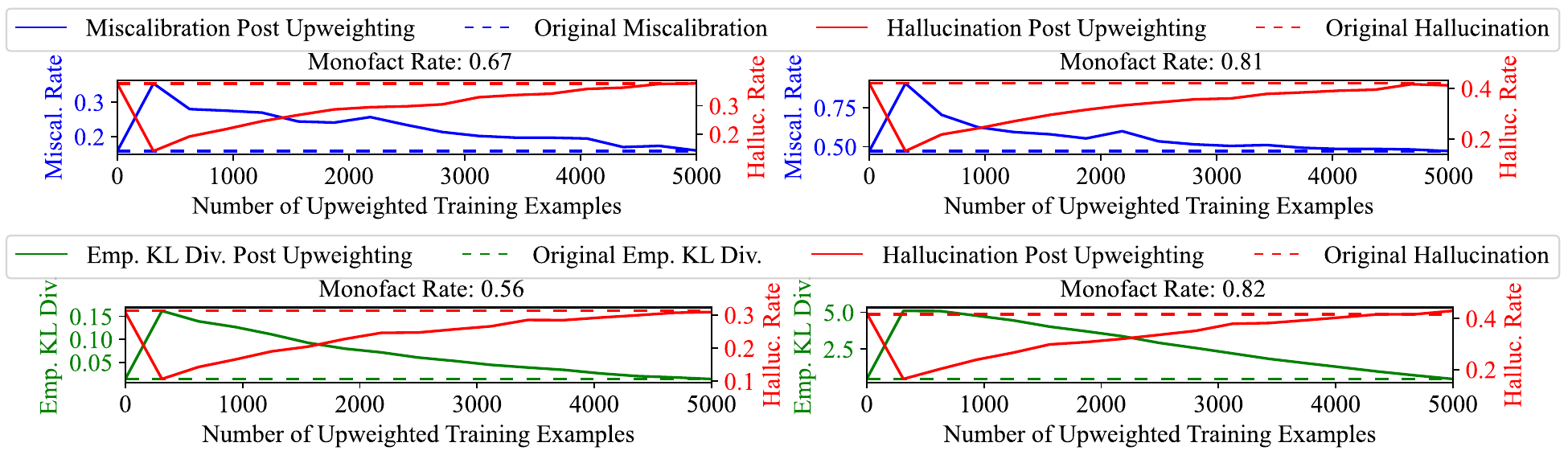}  
    \captionsetup{skip=4pt}
    \caption{\textbf{Top:} Relationship between miscalibration (blue line, left y-axis) and hallucination rates (red line, right y-axis) for select fixed monofact rates. Dotted lines indicate metrics prior to any interventions. Each subplot shows how miscalibration and hallucination evolve as we duplicate token occurrence for more and more statements from the training data (size of 5,000). \textbf{Initial miscalibration} (before upweighting intervention) is shown as the leftmost point of each blue curve at $k=0$ upweighted examples.
    \textbf{Bottom:} Relationship between empirical KL Divergence (green line, left y-axis) and hallucination rates (red line, right y-axis) for select fixed monofact rates.}
    \label{fig:upweighting_all}  
\end{figure*}

With monofact rates fixed, Fig.~\ref{fig:upweighting_all} traces the effect of the upweighting algorithm for controlled miscalibration injection. Upweighting just \(312\) examples (\(\approx\!6\%\) of the corpus) could significantly increase overall miscalibration and KL divergence while halving hallucination in some cases (Fig.~S4 in SI appendix shows additional results). Miscalibration and KL divergence return to its baseline when \(k\) nears \(|S|\) due to normalization. The hallucination reduction effect is most pronounced in high‑monofact situations. We find that our miscalibration injection method pushes bin-wise polarity even higher while increasing the total magnitude of miscalibration and KL divergence.

\section{Supervised Fine-tuning Methodology}
To test whether the theoretical relationship between monofact rates, miscalibration, and hallucination extends to modern language models, we fine-tuned Transformer models on synthetic biographical data with similarly controlled fact-frequency distributions. This approach allows us to systematically manipulate monofact rates while measuring hallucination in naturalistic text generation—a setting more representative of real-world applications than $n$-gram models.

We generated 10,000 unique biographies, each containing seven attributes: name, birth date, hometown, college, major, job title, and employer. Each attribute was sampled from pools of varying sizes to create natural frequency imbalances—for instance, 100,000 unique names versus 174 unique college majors—mirroring the long-tailed distributions observed in real-world knowledge bases. Biography text was assembled using four randomly selected template variants (e.g., ``[name] was born on [date]'' versus ``[name]'s birthday is [date]'') to introduce linguistic diversity while preserving factual content. Data pipeline details could be found in SI Appendix. 

Following recent work showing that power-law-distributed training data accelerates factual learning, we controlled monofact rates by sampling biographies according to Pareto distributions with varying shape parameters $\gamma$. Lower $\gamma$ values produce datasets where most biographies appear multiple times (low monofact rate), while higher $\gamma$ values approach uniform sampling where nearly every biography appears only once (high monofact rate). This generates training sets spanning monofact rates from 15.5\% to 99.8\% while maintaining a fixed vocabulary and total training budget.

 All models are fine-tuned using standard cross-entropy (negative log-likelihood) loss on next-token prediction:
\[
\mathcal{L} = -\frac{1}{|S|} \sum_{(x_1, \ldots, x_T) \in S} \sum_{t=1}^{T} 
\log p_\theta(x_t \mid x_{<t}),
\]
where $S$ is the training corpus and $\theta$ denotes model parameters. During upweighting, selected examples contribute to the loss multiple times per epoch according to their duplication factor.

\subsection{Upweighted Training}
\label{sec:LLM_method}

We conduct SFT traing on the \textsc{T5-Small}, \textsc{T5-Large} \cite{raffel2020exploring}, \textsc{GPT2-Medium}, and \textsc{GPT2-Large} \cite{radford2019language}, tracking four metrics throughout training: hallucination rate (measuring whether generated biographies contain correct attributes), miscalibration (total variation distance between $p$ and $g$ distributions), empirical KL divergence, and forced-generation inaccuracy (testing factual recall through prompted completion). This experimental design allows us to measure how data frequency distributions affect both what models learn and how confidently they express that knowledge. Specifically: 

\begin{itemize}[leftmargin=*,itemsep=2pt]
    \item \textbf{Inaccuracy}: The error rate on forced-generation factual recall. 
    Given a prompt (e.g., ``[Name]'s birthday is \underline{\hspace{1cm}}''), we measure 
    whether the model's most likely completion matches ground truth. This captures errors 
    during \emph{prompted recall}.
    \item \textbf{Hallucination rate}: $|G \cap F| / |G|$, where $G$ is the set of 
    generated statements and $F$ is the set of falsehoods. This captures errors during 
    \emph{free-form generation}, where the model produces plausible but false statements 
    unprompted.
\end{itemize}
These metrics are not complements: a model may achieve low inaccuracy on prompted recall 
while exhibiting high hallucination during open-ended generation, or vice versa.

\begin{figure*}[t]  
    \centering
    \includegraphics[width=0.7\textwidth]{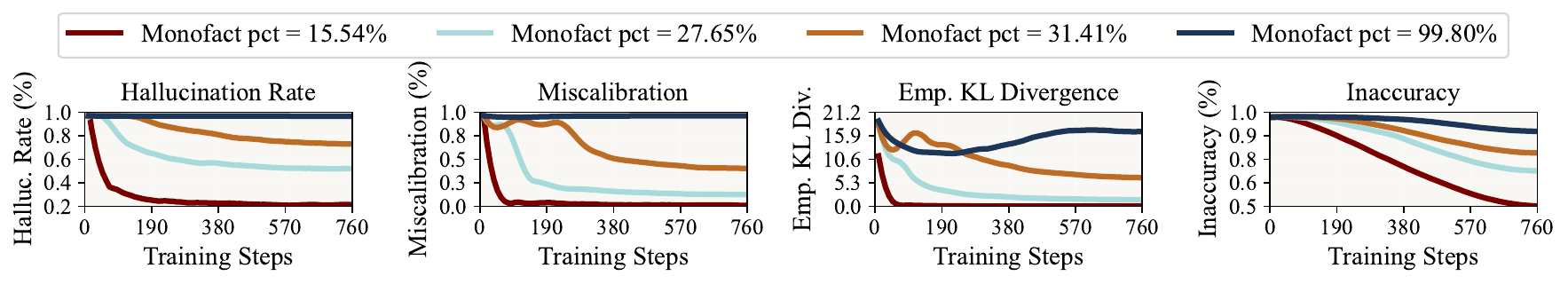}  
    \captionsetup{skip=4pt}
    \caption{(T5-Small) Four training statistics (from \textbf{Left} to \textbf{Right}: hallucination, miscalibration, KL divergence, inaccuracy) for every 5 steps during supervised fine-tuning for four training regimes with varying monofact rates but a fixed sample size of 10,000.}
    \label{fig:sft_linechart_noweight}  
\end{figure*}

\begin{figure*}[!h]  
    \centering
    \includegraphics[width=0.7\textwidth]{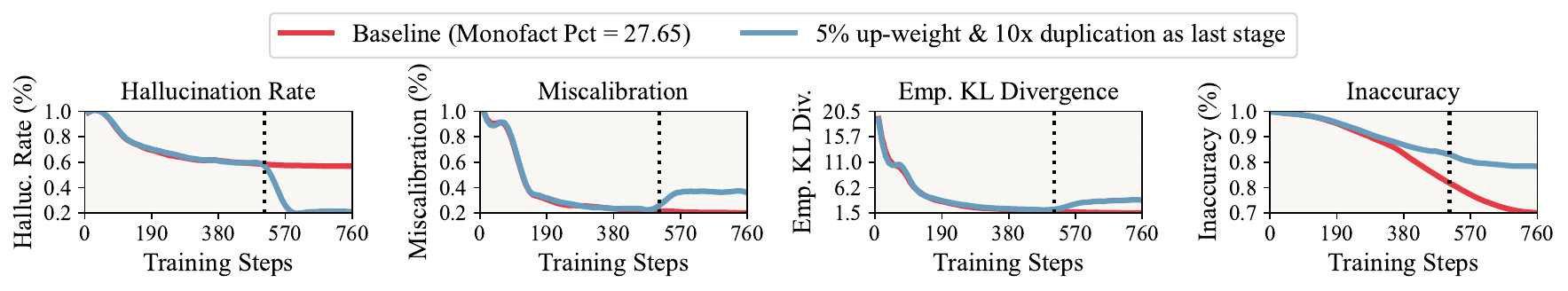}  
    \captionsetup{skip=4pt}
    \caption{(T5-Small) Four training statistics (from \textbf{Left} to \textbf{Right}): hallucination, miscalibration, KL divergence, inaccuracy) for every 5 steps during supervised fine-tuning by upweighting 5\% of the training data 10 times as the \textit{last} stage of fine-tuning for a training sample with 27.65\% monofact rate. Dashed black line indicates when \emph{last-stage} upweighting injection takes place. \textbf{Initial miscalibration} prior to intervention is visible as the value before the vertical 
dashed line.}
    \label{fig:sft_linechart_gamma2_last}  
\end{figure*}

\begin{figure*}[!t]  
    \centering
    \includegraphics[width=0.7\textwidth]
    {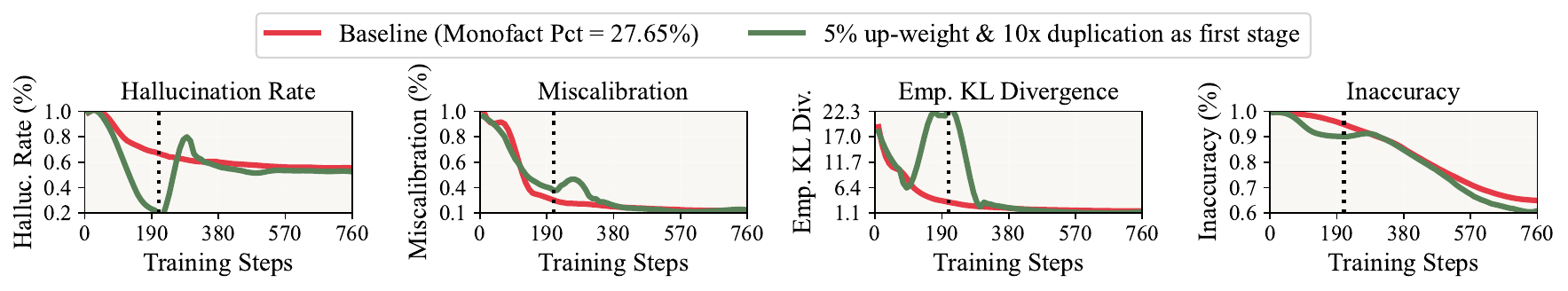}  
    \captionsetup{skip=4pt}
    \caption{(T5-Small) Four training statistics (from \textbf{Left} to \textbf{Right}): hallucination, miscalibration, KL divergence, inaccuracy) for every 5 steps during supervised fine-tuning by upweighting 5\% of the training data 10 times as the \textit{first} stage of fine-tuning for a training sample with 27.65\% monofact rate. Dashed black line indicates when \emph{first-stage} upweighting injection ends and normal SFT starts.}
    \label{fig:sft_linechart_gamma2_first}  
\end{figure*}

To test miscalibration injection as a hallucination mitigation strategy, we trained models under three upweighting regimes. The baseline regime trains without any sample reweighting, allowing us to measure the natural relationship between monofact rates and hallucination. In the last-stage upweighting regime, we train normally for the first two-thirds of training, then upweight a small fraction (5\%--30\%) of training examples by duplicating them multiple times (5$\times$ to 10$\times$) during the final training stage. This deliberately injects miscalibration after the model has learned most factual content. Conversely, the first-stage upweighting regime applies duplication early in training, then returns to standard training --- testing whether miscalibration injection works only as a final intervention or throughout learning. By comparing hallucination rates across these three regimes while holding monofact rates constant, we isolate the causal effect of intentional miscalibration on model outputs.

Additional information on training configurations, template specifications, and evaluation protocols appear in SI Appendix.

\section{Supervised Fine-tuning Results}
\subsection{Upweighted Training Results}
\label{sec: LLM_results}

In this section, we present detailed results for \textsc{T5-Small} and then discuss findings across additional model architectures and scales. Results are consistent across all four LLM models and model sizes unless otherwise noted.

Without any miscalibration injection, Fig.~\ref{fig:sft_linechart_noweight} illustrates for that low-monofact data facilitates faster convergence, hallucination prevention, and better overall accuracy, echoing the finding that data sampled from a Pareto distribution enables shorter knowledge acquisition plateaus~\cite{zucchet2025learn}.

Fig.~\ref{fig:sft_linechart_gamma2_last} tracks SFT metrics when miscalibration is injected at \emph{step 510} on a dataset with a  27.65\% monofact rate. Miscalibration and empirical KL divergence rise and remain elevated, yet the hallucination rate drops by roughly 40\%.  Overall inaccuracy dips slightly after the intervention and then stabilizes at roughly pre‑injection level, indicating that accuracy is not harmed by miscalibration injection. However, there is a clear hallucination-accuracy trade-off compared with the uninjected baseline: inaccuracy continues to inch downward while hallucination stays higher; the optimizer therefore favors overall accuracy improvements unless we intervene.

\subsection{Model Architecture Discussion}

To understand why last-stage upweighting reduces hallucination, we tested the same intervention at the start of training (Fig.~\ref{fig:sft_linechart_gamma2_first}) for Encoder-Decoder models \textsc{T5-Small} and \textsc{T5-Large}. Early-stage upweighting leaves terminal hallucination rates unchanged but reduces overall error slightly (1 - 3\%). We hypothesize that early injection aids initial learning of generic data representations, while late-stage injection reinforces previously encountered examples, thereby suppressing hallucination more effectively in encoder-decoder architectures. 

This pattern reverses for decoder-only models (Table.~\ref{tab:upweighting_timing}). Both \textsc{GPT2-Medium} and \textsc{GPT2-Large} benefit more from first-stage upweighting for hallucination mitigation, leading to reduction of ~8 - 10\%. This architectural dependence likely reflects differences in how bidirectional versus autoregressive attention mechanisms incorporate distributional biases during learning. Practitioners should therefore select upweighting timing based on their model architecture, with last-stage interventions for encoder-decoder models and first-stage interventions for decoder-only models.

Combining first and last-stage upweighting does not preserve the benefits observed from either intervention alone. Across all architectures tested, combined upweighting at both stages yields no improvement in hallucination rates, suggesting that excessive upweighting could disrupt the model's learned representations. 

\begin{table}[h]
\centering
\scalebox{0.87}{
\begin{tabular}{lccc}
\toprule
\textbf{Model Type} & \textbf{Last-stage} & \textbf{First-stage} & \textbf{First and Last-Stage} \\ 
\midrule
Encoder-Decoder (T5-Small) & $\downarrow$ & --  & -- \\
Encoder-Decoder (T5-Large) & $\downarrow$ & -- & -- \\
Decoder-Only (GPT2-Medium) & $\uparrow$ & $\downarrow$ & --\\
Decoder-Only (GPT2-Large) & $\uparrow$ & $\downarrow$ & --  \\
\bottomrule
\end{tabular}
}
\caption{Upweighting timing effects on hallucination across model architectures and scales. $\downarrow$ = reduces hallucination; $\uparrow$ = increases hallucination; -- = minimal to no impact.}
\label{tab:upweighting_timing}
\end{table}

\begin{figure*}[h]  
    \centering
    \includegraphics[width=0.85\textwidth]{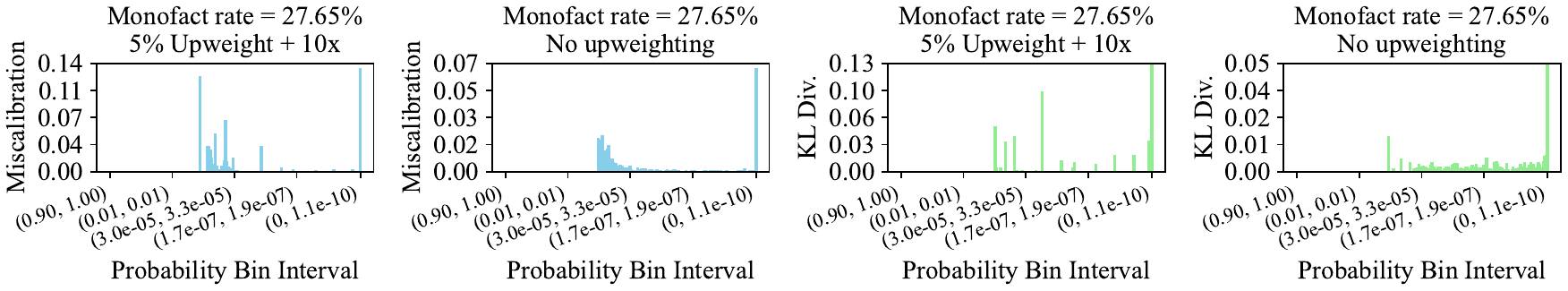}  
    \captionsetup{skip=4pt}
    \caption{(T5-Small) \textbf{Left:} Miscalibration per probability bin (binning created by a logarithmic binning strategy with $\epsilon$ = 0.1) for LLM SFT post 5\% upweighting with a duplication factor of 10. \textbf{Middle Left:} Average miscalibration per probability bin with no upweighting. \textbf{Middle Right:} Empirical kl divergence per probability bin post 5\% upweighting with a duplication factor of 10. \textbf{Right:} Empirical kl divergence per probability bin with no upweighting.}
    \label{fig:sft_upweighting_bin}    
\end{figure*}

\begin{figure*}[!h]  
    \centering
    \includegraphics[width=0.85\textwidth]{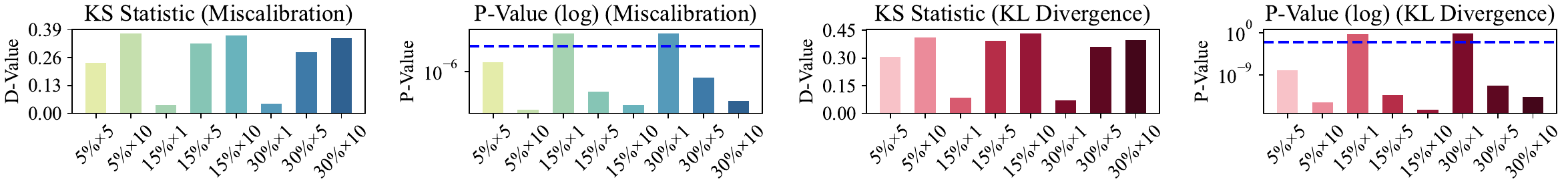} 
    \captionsetup{skip=4pt}
    \caption{Kolmogorov-Smirnov test statistics for pre and post injection bin-wise miscalibration \& KL divergence (monofact rate = 27.65\%). \textbf{Left and Middle Left:} D-values for 8 configurations, P-value results (log) for miscalibration. Blue dotted line represents sig. threshold of 0.01.  \textbf{Middle Right and Right:} KL divergence results. } 
    \label{fig:ks_test_stats}  
\end{figure*}

\subsection {Practical Guidance for Upweighting Interventions}
A key question remains: how do we decide on the magnitude and size of the injections? Fig. S2 in SI Appendix surveys a few upweighting configurations, formed by crossing three sample shares (5,15,30\%) with three duplication multipliers (1,5,10), under progressively higher monofact rates (15.5, 27.6, 31.4\%). 

To translate our empirical findings into actionable recommendations, we conducted two complementary analyses of upweighting configurations across varying monofact rates and computational budgets. First, a marginal utility analysis weights trade-offs between hallucination reduction, miscalibration increase, and computational cost according to different deployment priorities (e.g., hallucination-focused versus cost-conscious applications).

Across multiple scenarios, training data sampled with Pareto shape parameter $\gamma = 1.5$ (corresponding to ~27\% monofact rate) emerges as a robust choice, providing hallucination reduction (up to 38\%) with moderate miscalibration increases (up to 17\%) at 1.5$\times$ baseline computational cost. We also find that upweighting 15\% of training examples with 5$\times$ duplication during the final training stage offers the best cost-benefit ratio. For practitioners willing to accept higher computational costs, $\gamma = 2.0$ configurations on the Pareto frontier achieve hallucination reductions up to 62\% at 2.5$\times$ baseline cost. 

\subsection{Miscalibration Injection Results}

Fig.~\ref{fig:sft_upweighting_bin} shows that upweighting sharply redistributes probability mass toward more confident bins (left), yielding higher \emph{polarity}. To show that the impact of our injection on polarity is statistically significant, we run the Kolmogorov-Smirnov test (illustrated by Fig.~\ref{fig:ks_test_stats}) on bin-wise miscalibration and KL divergence distributions and find that all 8 upweighting configurations are statistically significant besides when duplication factor is 1, using a p-value threshold of 0.01. 


\section{Discussion}

Why does selective upweighting reduce hallucination? Our results suggest the mechanism operates through \emph{confidence 
redistribution} rather than simple memorization. By upweighting a subset of facts, we make the model overconfident on those specific instances, concentrating probability mass in high-confidence bins. This increased ``polarity'' 
reduces the probability of sampling from the uncertain tail of the distribution, where monofacts and plausible falsehoods are statistically indistinguishable. The model becomes more likely to generate facts it has learned with high confidence, rather than venturing into uncertain territory where hallucinations occur. In effect, the intervention strategically reshapes the confidence distribution to favor high-confidence generations over uncertain ones.

Our current intervention upweights a randomly selected subset of training examples. We hypothesize that targeted duplication of monofacts specifically would yield more effective hallucination reduction, particularly at high duplication factors. Under the Kalai-Vempala framework, monofacts are precisely the facts most vulnerable to hallucination: the model has seen them only once and assigns them lower confidence, placing them in probability bins where true and plausible-but-false completions are difficult to distinguish. Random upweighting may ``waste'' duplication effort on already-frequent facts that the model has learned with high confidence. Investigating targeted versus random upweighting is a promising direction for future work.

\section{Conclusion and Limitations}
\phantomsection
\label{sec:conclusion}

Hallucination is not a mysterious failure mode but a predictable outcome of monofact rates and model miscalibration. By unifying the Kalai–Vempala theoretical framework with controlled experiments, we demonstrate that training data sampled from heavy-tailed distributions naturally lowers monofact rates and reduces hallucination, and that intentional miscalibration --- injected through selective upweighting of training examples --- further decreases hallucination by up to 40\% without compromising accuracy. Our empirical KL-divergence bound provides a practical analog of the original theorem, requiring no knowledge of the true data distribution. The intervention is simple: reinforce a small, targeted subset of samples during the final stage of supervised fine-tuning. This reshapes calibration by concentrating probability mass in high-confidence bins while suppressing low-confidence predictions. These findings challenge the assumption that deduplication is universally beneficial and identify data-centric mechanisms for hallucination control.

Nevertheless, our approach has important limitations. Most significantly, selective upweighting may introduce unintended biases: models could become predisposed to generate upweighted facts during free generation, analogous to the "Golden Gate Claude" phenomenon where amplifying a single latent feature induced pervasive mentions of the Golden Gate Bridge \citep{templeton2024scaling}. Practitioners should therefore carefully audit which samples receive emphasis. Moreover, our experiments focus on structured biographical facts; generalization to other domains and fact types remains to be established. 

More broadly on limitations, our findings raise an important question about the tension between hallucination reduction and generalization. Standard deduplication practices exist primarily to prevent overfitting and promote generalization to unseen examples. While our results demonstrate that strategic duplication lowers hallucination rates, excessive duplication could impair the model's capacity for compositional reasoning or rule-based generalization. Consider arithmetic facts (e.g., ``$17 + 38 = 55$'') or letter-counting queries (e.g., ``How many R's are there in BLUEBERRY?''): these tasks require learning underlying rules rather than memorizing specific instances. Whether upweighting helps or harms such systematic generalization remains an open question for future investigation.

\acknow{We give very warm thanks to Adam Kalai, Santosh Vempala, Chris Callison-Burch, Lyle Ungar, and Bryan Li for helpful discussions.
}

\showacknow{}

\bibsplit[2]

\bibliography{pnas-sample}

\end{document}